\documentclass[sigconf]{acmart}

\usepackage{multirow}
\usepackage{graphicx}
\usepackage{url}
\usepackage{booktabs}
\usepackage{enumitem}
\usepackage{CJKutf8}
\usepackage{natbib}

\usepackage{color,xcolor}
\usepackage{enumitem}

\usepackage{colortbl}
\usepackage{graphicx}
\usepackage{subfigure}
\usepackage{soul}
\usepackage{url}
\usepackage{amsmath}
\usepackage[utf8]{inputenc}
\usepackage{xcolor}
\usepackage{algorithm}

\usepackage{amssymb}
\usepackage{makecell}
\usepackage{xspace}
\usepackage{array}
\usepackage[normalem]{ulem}
\usepackage{footmisc}
\usepackage{colortbl}
\usepackage{epstopdf}
\usepackage{setspace}
\usepackage{pifont}
\usepackage{balance}

\newcommand{\hlc}[2][yellow]{{%
    \colorlet{foo}{#1}%
    \sethlcolor{foo}\hl{#2}}%
}

\AtBeginDocument{%
  \providecommand\BibTeX{{%
    \normalfont B\kern-0.5em{\scshape i\kern-0.25em b}\kern-0.8em\TeX}}}

\copyrightyear{2024} 
\acmYear{2024} 
\setcopyright{acmlicensed}\acmConference[WSDM '24]{Proceedings of the 17th ACM International Conference on Web Search and Data Mining}{March 4--8, 2024}{Merida, Mexico}
\acmBooktitle{Proceedings of the 17th ACM International Conference on Web Search and Data Mining (WSDM '24), March 4--8, 2024, Merida, Mexico}
\acmPrice{15.00}
\acmDOI{10.1145/3616855.3635813}
\acmISBN{979-8-4007-0371-3/24/03}

\begin{document}

\begin{CJK*}{UTF8}{gkai}

%\title{Improve Retrieval-Based Chatbots via Dialogue Simulation: \\A General Framework for Multi-Turn Response Selection}
\title{Multi-Intent Attribute-Aware Text Matching in Searching}

\author{Mingzhe Li$^*$}
\affiliation{%
  \institution{Ant Group}
%   \country{dd}
}
\email{limingzhe.lmz@antgroup.com}

\author{Xiuying Chen*}
\affiliation{%
  \institution{CBRC, KAUST}
}
\email{xiuying.chen@kaust.edu.sa}

\author{Jing Xiang}
\affiliation{%
  \institution{Ant Group}
%   \country{dd}
}
\email{liping.xj@antgroup.com}

\author{Qishen Zhang$^{\dagger}$}
\affiliation{%
  \institution{Ant Group}
%   \country{dd}
}
\email{qishen.zqs@antgroup.com}

\author{Changsheng Ma}
\affiliation{%
  \institution{CBRC, KAUST}
}
\email{changsheng.ma@kaust.edu.sa}

\author{Chenchen Dai}
\affiliation{%
  \institution{Ant Group}
%   \country{dd}
}
\email{daichenchen.dcc@antgroup.com}

\author{Jinxiong Chang}
\affiliation{%
  \institution{Ant Group}
%   \country{dd}
}
\email{ jinxiong.cjx@antfin.com}

\author{Zhongyi Liu}
\affiliation{%
  \institution{Ant Group}
%   \country{dd}
}
\email{zhongyi.lzy@alibaba-inc.com}

\author{Guannan Zhang}
\affiliation{%
  \institution{Ant Group}
%   \country{dd}
}
\email{zgn138592@antgroup.com}

\def\authors{Mingzhe Li, Xiuying Chen, Jing Xiang, Qishen Zhang, Changsheng Ma, Chenchen Dai, Jinxiong Chang, Zhongyi Liu, Guannan Zhang}

\renewcommand{\shortauthors}{Mingzhe Li et al.}

\thanks{* Both authors contributed equally to this research.\\
$\dagger$ Corresponding author.} 
%%
%% The abstract is a short summary of the work to be presented in the
%% article.
\begin{abstract}
Text matching systems have become a fundamental service in most Searching platforms.
For instance, they are responsible for matching user queries to relevant candidate items, or rewriting the user-input query to a pre-selected high-performing one for a better search experience.
In practice, both the queries and items often contain multiple attributes, such as the category of the item and the location mentioned in the query, which represent condensed key information that is helpful for matching.
However, most of the existing works downplay the effectiveness of attributes by integrating them into text representations as supplementary information.
Hence, in this work, we focus on exploring the relationship between the attributes from two sides.
Since attributes from two ends are often not aligned in terms of number and type, we propose to exploit the benefit of attributes by multiple-intent modeling.
The intents extracted from attributes summarize the diverse needs of queries and provide rich content of items, which are more refined and abstract, and can be aligned for paired inputs.
Concretely, we propose a multi-intent attribute-aware matching model (MIM), which consists of three main components: \textit{attribute-aware encoder}, \textit{multi-intent modeling}, and \textit{intent-aware matching}.
In the \textit{attribute-aware encoder}, the text and attributes are weighted and processed through a scaled attention mechanism with regard to the attributes' importance.
Afterward, the \textit{multi-intent modeling} extracts intents from two ends and aligns them.
Herein, we come up with a distribution loss to ensure the learned intents are diverse but concentrated, and a kullback–leibler divergence loss that aligns the learned intents.
Finally, in the \textit{intent-aware matching}, the intents are evaluated by a self-supervised masking task, and then incorporated to output the final matching result.
Extensive experiments on three real-world datasets from different matching scenarios show that MIM significantly outperforms state-of-the-art matching baselines.
MIM is also tested by online A/B test, which brings significant improvements over three business metrics in query rewriting and query-item relevance tasks compared with the online baseline in Alipay App.

\begin{figure}
    \centering
    \includegraphics[scale=0.41]{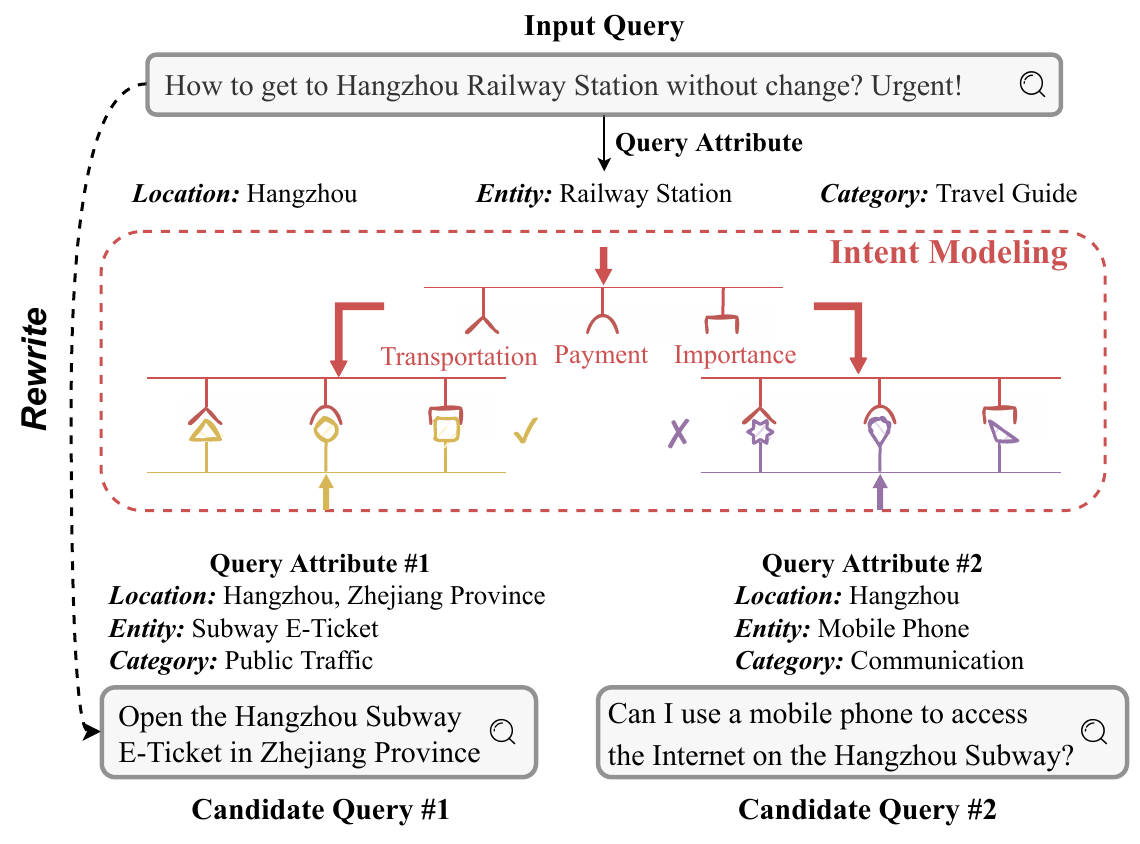}
    \caption{
        The application of our text matching task in query rewriting scenarios.
        The model rewrites the given query to pre-selected high-performing queries.
    }
    \label{fig:qqintro}
\end{figure}

\end{abstract}


% \ccsdesc[500]{Information systems~Web searching and information discovery}

%%
%% Keywords. The author(s) should pick words that accurately describe
%% the work being presented. Separate the keywords with commas.

%%
%% This command processes the author and affiliation and title
%% information and builds the first part of the formatted document.
\maketitle

\begin{figure*}
    \centering
    \includegraphics[scale=0.25]{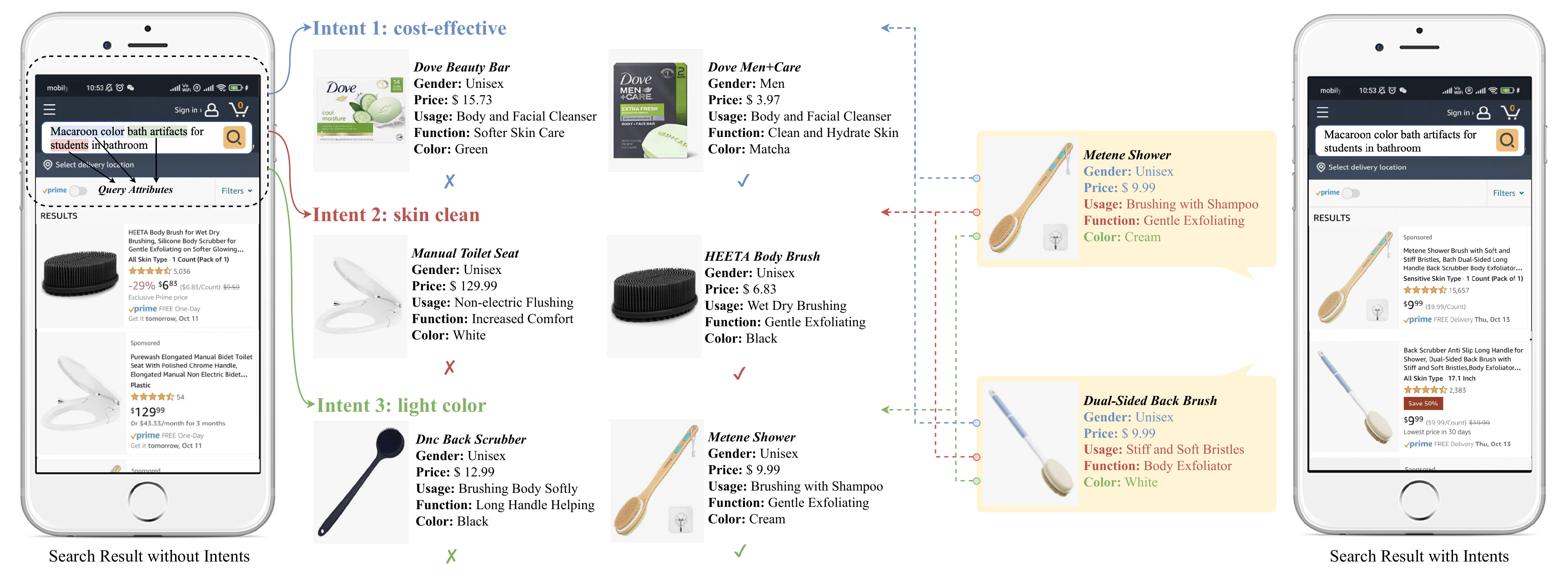}
    \caption{
        The application of our model in the query-item matching setting.
        The model identifies multiple intents on ``cost-effective'', ``skin clean'', and ``light color'', based on which it conducts query-item matching to return relevant items to the users.
    }
    \label{fig:intro}
\end{figure*}

\section{Introduction}
Search engines are one of the most important technologies nowadays.
Traditional search engines build inverted indexes based on queries and documents to accomplish information retrieval.
With the richness of search content, modern systems start to store key information of queries and items by attributes, such as brands, entities, and regions \cite{rashed2022carca}.
Such information can help classify whether the queries or items match each other.
For example, in the query rewriting task in Figure~\ref{fig:qqintro},
when a user inputs a query, the matching system can identify the related attributes, and rewrite the query to high-performing queries in a curated set, so as to increase the chance of matching relevant items further.
% \textcolor{black}{when a user inputs a query, the retrieval system first retrieves a set of pre-selected high-performing queries that are potentially relevant through multi-stage recall. Then, the matching system identifies the related attributes, and filters out any erroneous rewrites, so as to increase the chance of matching relevant items further.}
Another application is the query-item matching task in Figure~\ref{fig:intro}, where the system matches the query to related items, also based on the attributes.

Various approaches have been adopted for text matching with attributes in search platforms \cite{sarwar2001item,li2022keywords,trabelsi2022strubert}.
Recently, deep-learning-based approaches become popular, which typically represent queries and items with low-dimensional vectors.
% To name a few, \cite{chen2017cross} fused food attributes, particularly the interplay among ingredients, cutting, and cooking methods, into the text-based recipe representation for cross-modal recipe retrieval.
% % \cite{chen2017cross,trabelsi2019improved} presented a context-aware table retrieval method that is based on embeddings for attribute tokens.
% \cite{trabelsi2019improved} presented a context-aware table retrieval method based on augmenting contextual representation with attribute tokens.
% Most recently, \cite{kong2022multi} proposed to explicitly represent multiple attributes using one embedding per aspect and fused the attribute embeddings for representing queries and documents.
To name a few, \cite{zou2022divide} extracted keywords and intents from the input texts as attributes, and enhances the matching from different granularities.
\cite{kong2022multi} treated multiple additional inputs as attributes, and explicitly represented each attribute as an embedding, which is further fused into the input embeddings.
Most recently, \cite{shan2023beyond} considered the relationship between attributes and input text by introducing an attribute fusion layer to identify the most relevant features.

The above works show that attributes play an important role in text matching task.
However, they mainly integrate the attribute information into the text representation, underplaying the effectiveness of attributes.
Since attributes contain a condensed form of information in the input, we assume that modeling the relationship between the attributes of two sides is decisive in the matching tasks.
Nevertheless, attributes from two ends are often not aligned in terms of number and type, and directly matching them is counterintuitive.
We give a real-world example here in Figure~\ref{fig:intro}.
The query ``Macaroon color bath artifacts for students in bathroom'' with attributes ``students'', ``bath artifacts'' and ``macaroon color'' can be matched with the \textit{back rub}, which has \textit{a low price} fit for a student and \textit{light color} that belongs to the macaroon color scheme.
However, the number of attributes for the items is different from that of the query, and the attribute content does not directly correspond to the query either.
Thus, a deeper and more comprehensive understanding of the attributes is required.
In this paper, we resort to the ``intents'' of the text, which is a more abstract and representative concept that reflects the customers' needs.
A suitable rewriting query or an item will always meet the customers' intents.
In the above case, once the model understands the query by its intents on ``cost-effective'', ``light color'', and ``skin clean'', it can match the query to the proper items.

%模型部分没问题。
Motivated by the above observation, in this paper, a novel multi-intent attribute-aware matching framework (MIM) is proposed. 
Specifically, MIM consists of three main components: \textit{attribute-aware encoder}, \textit{multi-intent modeling}, and \textit{intent-aware matching}.
% Herein, we use the matching between a query and an item as an example to introduce our model at a high level.
Firstly, an attribute-aware encoder processes the query and attributes from two inputs together, to let information flow across both parts.
Considering that different attributes have different impacts on the matching performance, we propose a scale attention mechanism and assign different weights to the attributes in the encoding process.
Afterward, a multi-intent modeling module extracts a number of intent representations from the attribute-aware representations, where each intent reflects a different intention of the user.
To ensure the learned intents can represent diverse perspectives of the input text while also closely related to the text gist, we come up with a distribution loss.
This loss pushes the intent representations far away from each other but centered around the text representation.
We further propose a kullback-leibler loss to align the intent distributions and match the learned intents from two sides.
Finally, an intent-aware matching module takes the learned intent and text representations as input to give the final matching result.
Since some salient intents can be decisive in the matching evaluation, while other second-order intents may play a less important role, we come up with a self-supervision task to evaluate the importance of each intent.
Concretely, we mask each intent and see how the change influences the objective loss.
If the performance gets worse, which means that the intent is important for the final decision, we increase its corresponding weight and vice versa.
Offline experiments on three real-world datasets in different scenarios, and online A/B test experiments on Alipay App verify the effectiveness of our proposed MIM.

To summarize, our contributions are three-fold:

$\bullet$ We propose MIM for multi-intent text matching learning in searching systems, which is the first attempt to comprehensively leverage both query and item textual attributes for multi-intent understanding.

$\bullet$ To achieve the above goal, we propose an intent-aware encoder, a multi-intent modeling module, and an intent-aware matching module, which extract multiple intents from the queries and the matched intents from the items to perform precise and diverse candidate matching.

% To ensure the extracted intents are concentrated, diverse, and accurate for the matching framework, we devise a distribution loss and a KL loss, which pushes the intent representations far from each other and centered around the input text, and aligns the intents from both sides, respectively.

$\bullet$ Extensive experiments on three real-world applications validate the effectiveness of the proposed method. 
Currently, MIM has been deployed online, serving hundreds of millions of Alipay users and yielding significant improvement in commercial metrics.

\section{Related Work}
\label{sec:related}
% Our current research is built upon previous work in three fields: e-commerce item relevance searching, query rewriting, and multi-intent modeling.

\nocite{liang2023let,li2021stylized}
\noindent \textbf{E-Commerce Item Relevance Searching.}
Millions of items can be found on typical e-commerce platforms.
Users go to these platforms and type in search terms to find the things they want.
The success of e-commerce platforms depends on matching the query to the appropriate items.
Users visit these platforms and enter search queries to retrieve their desired items~\cite{liu2022category,kong2022multi}.
Therefore, matching the query to the relevant items is essential for the success of e-commerce platforms~\cite{fan2022modeling}. 
To address the challenge of semantic alignment between queries and items, several works attempted to incorporate contextual information~\cite{kumar2021neural, thakur2021augmented} to obtain more comprehensive representations.
To better learn the relative position of sentence vectors in the semantic space, \cite{gao2021simcse, yan2021consert, carlsson2020semantic} provided difficult negative samples for contrastive learning through optimizing the sampling strategy.
These works generally focused only on the pure input text of query and item, neglecting the additional attribute information along with the text, which is proved to be useful in e-commerce tasks~\cite{kong2022multi,trabelsi2022strubert}.
\cite{zou2022divide} extracted keywords and abstract intents from sentences and performed semantic matching at different granularities.
\cite{shan2023beyond} modeled the relationship between attributes and input text by introducing an attribute fusion layer to identify the most relevant features.
% integrated various attribute information into the text representation by using multi-task learning.
However, previous works have undervalued the effectiveness of attributes by treating them as supplementary information integrated into text representations.
Therefore, in this work, we concentrate on exploring the relationship between attributes from two sides directly.

\noindent \textbf{Query Rewriting}
Aside from matching queries to items, another application of text matching is to rewrite user queries to pre-selected high-performing queries.
The user query might be of poor quality due to colloquial expression, errors in the input system, or the users' abridged language~\cite{zhou2021learned}.
The problem of vocabulary gap is also pronounced in the e-commerce setting where queries are in informal language whereas the item titles/descriptions are written in formal language \cite{kumar2021neural}.
% A recent deep learning approach \cite{he2016learning} leveraged the query-item bipartite graph built from click-stream data to rewrite/match a poorly performing query to a well-performing query.
% \cite{ponnusamy2020feedback} proposed a Markov chain model as a collaborative filtering mechanism to mine users’ reformulation patterns.
% If a query is an exact text match with a defect query, a rewrite will be triggered using the offline mined pairs.
% Different from the Markov chain model, \cite{chen2020pre} proposed a retrieval model for query rewriting tasks, utilizing pretrained language model as encoder.
% Query rewriting problem has been explored for defect reduction in question-answering task \cite{bonadiman2019large, yu2020few, liu2021conversational}.
\textcolor{black}{
% Query rewriting aims to transform these queries into more accurate and standard forms through semantic transmission.
\cite{miao2019cgmh} proposed to learn a transition probability to delete or replace words at the granularity level.
\cite{chen2020rpm} utilized abstracted keywords to modify the semantic representation of the input text.
\cite{qiu2021query} formulated query rewriting into a cyclic machine translation problem to leverage abundant click log data.
Further, in this paper, not only the attribute information is differentially incorporated into the semantic information, but also the alignment relationship between the two input texts at different levels of granularity is learned by extracting the intent embedded in the attributes.
}
% However, previous works did not take into account the correspondence between different attributes in the query. 
% In this paper, we further investigate the application of attributes in matching.

\noindent \textbf{Multi-Intent Modeling.}
Due to the diversity of the text and multiple matching possibilities in the text-matching task, multi-intent based encoding methods have been explored for better matching performance~\cite{cen2020controllable,li2020vmsmo,chen2023topic}.
For example, \cite{chai2022user} extracted multi-interests from the user's historical behaviors and profile information to enhance candidate matching performance in the recommendation system.
\cite{li2022miner} applied the multi-intent ideology in news recommendation and proposed a poly attention scheme to learn multiple interest vectors for each user, which encodes the different intentions of the user.
A variation of multi-intent, i.e., multi-view modeling, has also been explored.
\cite{zhang2022multi} proposed a multi-view document representation learning framework since a document can usually answer multiple potential queries from different views.
% Multi-view has also been tested in other tasks such as knowledge graph completion.
\cite{niu2022cake} proposed a multi-view link prediction task to determine the entity candidates that belong to the correct concepts in the commonsense view and face view.
However, to the best of our knowledge, no works explore the multi-intent phenomenon in the text matching field in Searching.
In this work, we come up with a multi-intent modeling module by aggregating information from text and attributes.

\begin{figure*}
    \centering
    \includegraphics[scale=0.35]{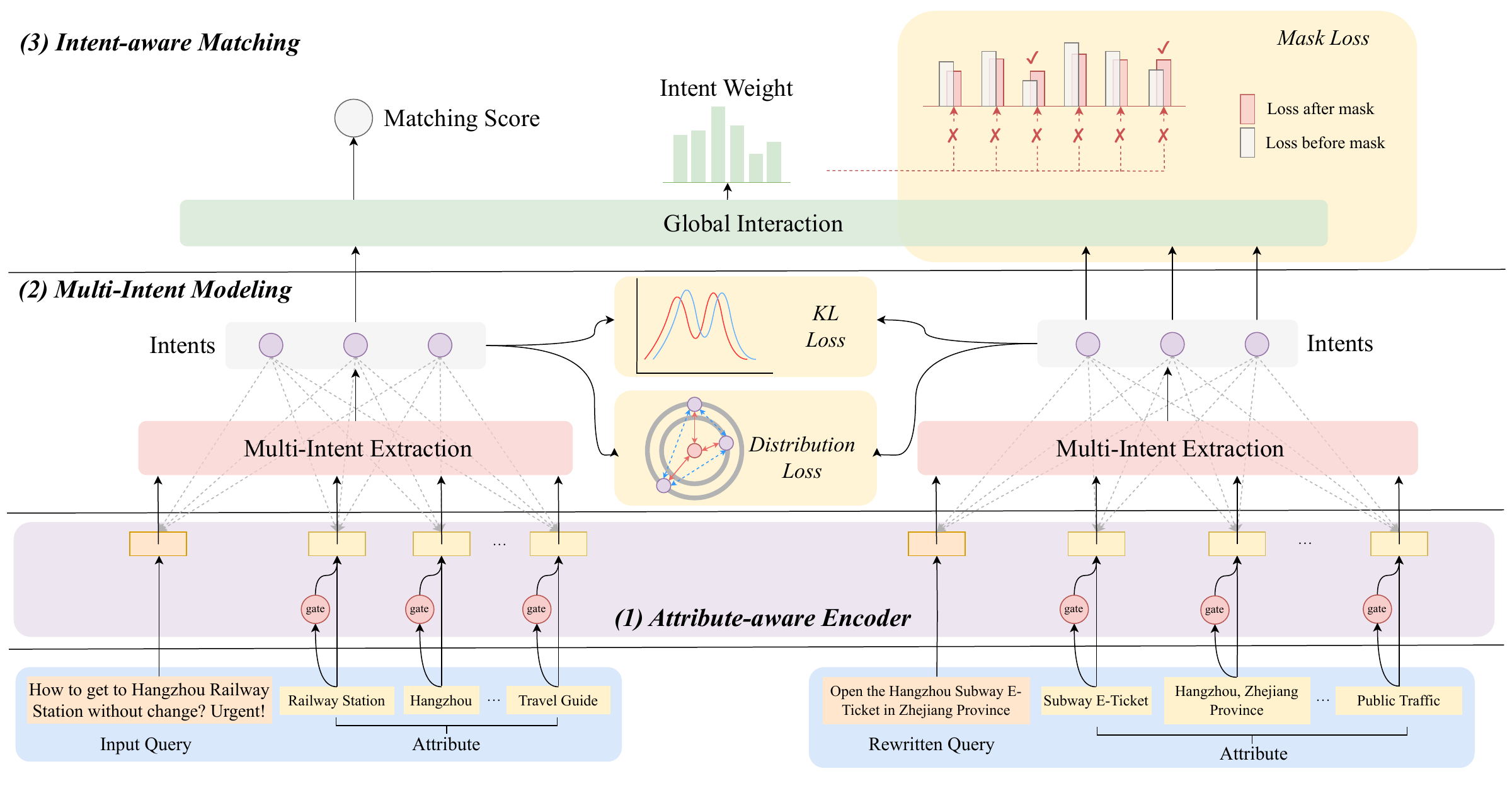}
    \caption{
        Overview of MIM, which consists of three parts: (1) \textit{Attribute-aware Encoder} obtains vector representations for the queries and attributes based on cross-attention and attribute-weighting; (2) \textit{Multi-Intent Modeling} extracts multiple diverse intents related to the queries with two devised losses; (3) \textit{Intent-Aware Matching} incorporates information from queries and intents to output the final matching result.
    }
    \label{fig:overview}
\end{figure*}

\section{Model}
\subsection{Problem Formulation}
\label{sec:formulation}

% Before presenting our approach for the attribute-aware text-matching task, 

We first introduce the notations and key concepts.
For brevity, we take the scenario of matching a user query to a pre-defined query as an example, and the settings for query-item matching is similar.
For an input query text $X=\{x_1, x_2, \dots, x_m\}$, where $x_i$ represents the $i$-th word, we assume there are corresponding attributes with the query such as entities, location area, and category words.
% These attributes are obtained by exact string matching or classification modules in practice.
We denote the attributes as $A=\{a_1,a_2,\dots,a_{n_A}\}$, where $n_A$ is the attribute number of $A$, and $a_i$ consists of $l_i^a$ words $\{a_i^1,a_i^2,\dots,a_i^{l_i^a}\}$.
% Here we take the attribute number as 3 since our online query rewrite system consists of 3 attributes.
Our aim is to identify whether the input query matches a rephrased query $Y=\{y_1, y_2, \dots, y_n\}$, where $y_j$ represents the $j$-th word.
The rephrased query $Y$ is also attached with $n_B$ attributes $B=\{b_1,b_2,\dots,b_{n_B}\}$.
In our multi-intent text matching model, we first extract multiple intents from the text and attributes, and aggregate information from the input with intents to generate the final matching score.

%\section{Model}
% \label{sec:model}
% \subsection{System Overview}
% In this section, we propose our multi-intent attribute-aware matching model (MIM), which is made up of three components as shown in Figure~\ref{fig:overview}:

% $\bullet$ \textit{Attribute-aware Encoder} obtains vector representations for the queries and attributes based on self-attention between multiple inputs and attribute-weighting, considering the importance of each attribute in regard to the query.

% $\bullet$ \textit{Multi-Intent Modeling} extracts multiple intents not only consisting of diverse interests but also related to the queries with a distribution loss.
% Then it aligns the learned intents with the kullback–leibler divergence loss.
% % with two devised losses,
% % e.g., the distribution loss and kullback–leibler divergence loss, respectively.

% $\bullet$ \textit{Intent-aware Matching} incorporates information from queries and intents to output the final matching result.
% The importance of each intent is decided based on a self-supervised mask loss.

% Note that in the following part we will also take query-to-query matching during query rewriting as an example.

\subsection{Attribute-aware Encoder}
\label{sec:dialog_generator}
First, we need to learn the semantic meaning of the input text and attributes from both queries.
% Attention mechanisms have become an integral part of compelling sequence modeling in various tasks~\cite{bahdanau2014neural,gao2020learning,shan2023beyond}, and in our text-matching task,
% % we also need to let words fully interact with each other to model the dependencies of words without regard to their locations in the inputs.
%  we concatenate the two queries with the attributes for the attention mechanism.
Considering the superior performance of attention mechanisms on text-matching task~\cite{bahdanau2014neural,wang2023towards}, we concatenate the input texts with the attributes.
% In the traditional attention mechanism in Transformer~\cite{vaswani2017attention}, the importance of each input attribute is calculated based on its relationship with other input units.
% However, we notice that the importance of each attribute also depends on its own content.
However, we notice that the importance of attributes is not only derived from their relationships with other input units, but also influenced by their own content.
If the attribute does not convey meaningful information related to the matching, it should play a less important role in the encoding process.
Hence, we propose a scaled self-attention module to model the temporal interactions between the texts and attributes in a matching pair.
% Generally, we calculate a gate for each attribute to decide its importance and fuse the gate information into the attention process.
% Hence, we assume that the importance of each attribute should also be decided based on its content.

% We first introduce how to obtain the gate value, and then introduce the fusing process.
Concretely, we append a special token ``$[CLS]$'' at the beginning of the concatenated input, and append a special token ``$[SEP]$'' to the front of each attribute and text.
The representation of ``$[SEP]$'' appended to the $k$-th attribute is regarded as the overall representation of the attribute, denoted as ${h}^{Attr}_k$.
Then, we obtain the $k$-th importance gate $g_k$ based on ${h}^{Attr}_k$: $g_k=\sigma({h}^{Attr}_k)$,
% \begin{align}
%     g_k=\sigma({h}^{Attr}_k), \quad k \in [1,n_A+n_B],
%     \label{equ:gate}
% \end{align}
where $k \in [1,n_A+n_B]$, and $\sigma(.)$ is a fully-connected layer with the sigmoid function.
% Note that there can be multiple words in the $k$-th attribute, and all the words correspond to the same importance gate $g_k$.
% For brevity, for the following part, we change the index of the gate from the attribute index to the word-level index.
% Considering the significance of the query text, we set its gate value $g_0$ as 1.
Next, we incorporate the gate values into the attention process.
Note that there can be multiple words in the $k$-th attribute, and all the words correspond to the same importance gate $g_k$.
For brevity, for the following part, we change the index of the gate from the attribute index to the word-level index.
Following \cite{vaswani2017attention}, all inputs are transformed into query $Q$, key $K$, and value $V$ through fully connected layers.
% The attentive module in the Transformer has three inputs: the query $Q$, the key $K$, and the value $V$.
% We use three fully-connected layers with different parameters to project the representations of texts or attributes into three spaces:
% \begin{align}
%     Q_i=FC(h^o_i),\ 
%     K_i=FC(h^o_i),\ 
%     V_i=FC(h^o_i),
% \end{align}
% where $h^o_i$ is the output representation of the previous layer.
% Due to the similar structure of each layer, we omit the superscript $l$ here.
The attentive module then takes each $Q_i$ to attend to $K_j$ as $\alpha'_{i,j} = Q_i \times K_j$.
Next, the gate score $g_j$ is incorporated in the gate-aware softmax operation, obtaining attention distribution $\alpha_{i,j}$ following $\alpha_{i,j} = \frac{\exp\left( g_j \cdot \alpha'_{i,j}\right)}{\sum_{n=1}^{L} \exp\left(g_j \cdot \alpha'_{i,n}\right)}$, where $\alpha_{i,j}$ denotes the attention weight between $i$-th word to $j$-th word, and $g_j$ denotes the corresponding attribute weight of the $j$-th word.
These output attention distribution results are used as weights to gain the weighted sum of $V_j$ as $h'_i = \textstyle \sum_{j=1}^{L} \alpha_{i,j} \times V_j$.
Here, the concatenated sequence length is calculated as $L=m+\sum\nolimits_{k=1}^{n_A} l_k^a+n+\sum\nolimits_{k=1}^{n_B} l_k^b$.
% shown in Equation~\ref{equ:transformer-sum}:
% % Next, we add the original input word representations of this layer $h^o_i$ on $h'_i$ as the residual connection layer, shown in Equation~\ref{equ:drop-add}:
% \begin{align}
%     \alpha'_{i,j} &= Q_i \times K_j, \label{equ:attention}\\
%     \alpha_{i,j} &= \frac{\exp\left( g_j \cdot \alpha'_{i,j}\right)}{\sum_{n=1}^{L} \exp\left(g_j \cdot \alpha'_{i,n}\right)}, \label{equ:gate_attention}\\
%     h'_i &= \textstyle \sum_{j=1}^{L} \alpha_{i,j} \times V_j, \label{equ:transformer-sum}
%     % \hat{h}_i &= \text{Dropout} \left( h^o_i + FC(h'_i) \right), \label{equ:drop-add}
% \end{align}
% \textcolor{black}{
% where $\alpha_{i,j}$ denotes the attention weight between $i$-th word to $j$-th word, the concatenated sequence length $L=m+\sum\nolimits_{k=1}^{n_A} l_k^a+n+\sum\nolimits_{k=1}^{n_B} l_k^b$, and $g_j$ denotes the corresponding attribute weight of the $j$-th word.
% }
To prevent the vanishing or exploding of gradients, a layer normalization operation~\cite{lei2016layer} is also applied on the output of the feed-forward layer.
% as shown in Equation~\ref{equ:ffn}: 
% \begin{align}
%     % h_i &= \text{norm}(FC(\max(0, FC(\hat{h}_i)))), 
%     h_i &= \text{norm}(FC(h^o_i + FC(h'_i))),
%     \label{equ:ffn}
% \end{align}
% where $h_i$ denotes the output hidden state of the $i$-th word in the Attribute-aware Encoder.
% where $W_1, W_2, b_1, b_2$ are all trainable parameters, and $h_i$ denotes the output hidden state of $i$-th word in the Transformer.

% We denote the final representations of the $i$-th attribute as $h^A_i$ and $h^B_i$ for $A$ and $B$, and queries $X$ and $Y$ as $h^X$ and $h^Y$,  respectively.

\textcolor{black}{
Herein, we use the corresponding ``[SEP]'' vector for the $i$-th attribute at the last layer as the updated attribute representations $h^A_i$ and $h^B_i$ for $A$ and $B$ respectively.
}
% As for encoding the input query $X$, we also extract the vector of ``[SEP]'' at the first position of text $X$ as its entire representation, denoted as $h^{cls}_X$, which is similar in rewritten query $Y$ and denoted as $h^{cls}_Y$.

% \begin{figure}
%     \centering
%     \includegraphics[scale=0.56]{figs/encoder.pdf}
%     \caption{
%         The Attribute-aware Encoder structure of our proposed MIM.
%     }
%     \label{fig:encoder}
% \end{figure}

\subsection{Multi-Intent Modeling}
\label{sec:knowledge selector}
\textbf{Multiple Intent Extractor.}
After obtaining the vector representations of the inputs, we aim to analyze and extract the multiple intents from them.
The motivation is twofold.
Firstly, taking Figure~\ref{fig:qqintro} as an example, a query may reflect the user's multiple intents in regard to a candidate rewriting query, e.g., ``transportation'', ``payment'', and ``importance''.
On the other hand, two queries can be similar in terms of word usage but have different semantic meanings.
Without intent modeling, it would be hard to identify the subtle differences and give correct matching results.
For example, ``How to use mobile phones to take the Hangzhou Subway?'' and ``Can I use a mobile phone to access the Internet on the Hangzhou Subway?'' should not be matched together, since they have different user intents.
Hence, if we can extract the hidden intents behind the input and evaluate the matching degree between two queries taking the intents into consideration, the matching will be more comprehensive and accurate.
% In this subsection, we aim to abstract and summarize the semantic meanings of the inputs into intents, and in the next subsection we utilize them to perform matching. 

Generally, we aggregate attributes into multiple intents based on the association between them.
Considering that the query is playing a central role during the intent extraction, we treat it as a condition, which controls the intent extraction process.
Concretely, $h^X$ is concatenated with each updated attribute representation $h^A_j$, and the multi-intent can be obtained through attribute-weighted aggregation:
\begin{align}
    w_{c,j} &= \text{softmax}([h^X;h^A_j]W^A+b^A),\\
    I^X_c &= w_{c,j}^\text{T}\times h^A_j,
\end{align}
where $[;]$ denotes concatenation operation, and $W^A \in \mathbb{R}^{2d \times c}$ and $b^A\in \mathbb{R}^{c}$ are learnable matrices.
The weight $w_{c,j}$ aggregates $n_A$ attributes into $c$ intents.
In this way, we obtain the multi-intent $I^X \in \mathbb{R}^{c \times d}$ of input query $X$, where $d$ is the hidden dimension.
The intent extraction of rewritten query $Y$ with attributes $B$ is similar, and we denote it as $I^Y$.

% Concretely, for each query, we calculate initial intent representations by the weighted sum of the attributes, where the weights are determined by the text:
% \begin{align}
%     I_i &= FC(a). \label{equ:ffn}
% \end{align}

\textbf{Distribution Loss.}
Intuitively, we want the learned intents to be diverse and conclude all the interests implied by the query.
However, multiple intents can usually degrade into one, as shown in similar one-to-many architectures such as the mixture of experts \cite{zuomoebert} and multi-view matching \cite{zhang2022multi}.
On the other hand, the semantic meaning of learned intents should be closely related to that of the user query.
Hence, we come up with a distribution loss, which pushes the representations of multiple intents to be far from each other, but centered around the query representation.
Concretely, we calculate the similarity between each pair of intents and between the query and intent.
The distribution loss minimizes the similarity between intents and their distance to the query representation:
\begin{align}
    \mathcal{L}_{dis} = \mathbb{E}[-\text{log}\frac{e^{cos(I_i, h^*) / \tau}}{\sum_{I_i,I_j \in I^*}e^{cos(I_i, I_j)/\tau}}],
\end{align}
where $\tau$ denotes the temperature parameter, $*$ can be $X$ or $Y$.
$I_i, I_j \in I^*$ is the extracted intents, $cos(.)$ denotes cosine similarity between the two representations, and $h^*$ denotes the query representations from the attribute-aware encoder.
% To the best of our knowledge, this is the first work that explores multi-intent learning in an e-commerce text-matching task.

\textbf{KL-Divergence Loss.}
Apart from the distribution restriction on the process of learning intents, we also have a divergence loss on the learned intent representations. 
% Since we have multiple intents from each input side, they form two distributions.
The motivation is that for paired queries, their intents in the latent space should also be aligned with each other.
% Otherwise, their intents should not match each other.
The intent matching is especially important when the attributes cannot be directly aligned.
Hence, for the positive cases in our training dataset, we come up with the intent KL-divergence loss to push their intent representations to have similar distributions.
On the other hand, for the negative pairs in the training set, we aim to maximize the KL-divergence.
We choose to align the intent distributions instead of directly pushing the intent representations closer since we emphasize the ``matching'' of the multiple representations: 
\begin{align}
    \mathcal{L}_{KL}&=\mathcal{D}_{K L}\left(q(I^X \mid h^X, h^A_i) \| q(I^Y \mid h^Y, h^B_i)\right),
    % \mathcal{L}_{KL}&=\frac{1}{2}(\mathcal{D}_{K L}\left(\mathcal{P}_1^w\left(y_i \mid x_i\right) \| \mathcal{P}_2^w\left(y_i \mid x_i\right)\right)\\&+\mathcal{D}_{K L}\left(\mathcal{P}_2^w\left(y_i \mid x_i\right) \| \mathcal{P}_1^w\left(y_i \mid x_i\right)\right)).
\end{align}
where $q$ denotes the statistic distribution of the generated intents.

\subsection{Intent-aware Matching}
After obtaining the multi-intent representation, we can utilize them to help predict the matching degree of the two inputs.
% Our matching mechanism consists of a local and a global interaction.
% Firstly, a self-attention \textcolor{black}{similar as Transformer~\cite{vaswani2017attention}} is adopted among intents $I^X$ and $I^Y$ of each input for local interaction.
% % and utilizes the similarities between multi-intent to update its representation.
% The intent vectors thus become $\hat{I}^X$ and $\hat{I}^Y$.

% Next, we let information fully flow between the two inputs.
% To begin with, the two inputs are concatenated as $I_j$, where $j \in [1,2c]$.

Concretely, we incorporate the intent information into text representations by utilizing attention mechanisms.
Remember that in \S~\ref{sec:dialog_generator}, we append a ``[CLS]'' token in front of both inputs.
% In order to distinguish the importance of different intents,
% In order to select the relevant intents from both sides,
Here, we apply its polished representation $h^{cls}$ output from the attribute-aware encoder as a query, and concatenate the intents $I^X$ and $I^Y$ of each input as $I_j$, so as to calculate the intent importance weight $\beta_{j}$ during the matching:
\begin{align}
    \beta_{j}&=\frac{\exp\left(h^{cls} \times I_j \right)}{\sum_{t=1}^{2c} \exp\left(h^{cls} \times I_t\right)},
\end{align}
where $c$ is the extracted intent number of each input.
Here, the global information from ``[CLS]'' token is fused into the matching process.
Finally, the likelihood that the input query $X$ should be rephrased as rewritten query $Y$ can be obtained as:
\begin{align}
    P=\sigma(FC([h^{cls};\beta_1 \hat{I}_1;\dots;\beta_j \hat{I}_j])),
\end{align}
where $\sigma$ is the sigmoid function.

The model training is then to maximize this likelihood score for each positive target item against the rest negative ones:
\begin{align}
    \mathcal{L}_{match}=-(s\log(P)+(1-s)\log(1-P)),
\end{align}
where $s$ denotes that inputs match (1) or not (0).

\textbf{Intent-Mask Self-Supervision Task.}
Motivated by the self-supervision training from the existing pretrained language models \cite{liu2019roberta,lewis2020bart}, we propose an intent-mask self-supervision task to identify intents that are beneficial for the matching and emphasize their contributions in the matching process.
% The role of each attribute varies depending on its content, while the importance of each intent is also various depending on the compared queries.
% For example, in the case shown in Figure~\ref{fig:intro}, when matching the query to the first candidate rewriting query ``cheap bath tools'', the ``cost-effective'' intent is more important compared with ``light color'' intent.
% Herein, we propose a self-supervised task to help identify the more important intents and emphasize their contributions in the matching process.
% Generally, we iteratively mask each intent and calculate the new target loss $L_{new}$ in such a case.
% If the origin loss $L_{old}$ increases by a large margin, it means that the masked intent is of great importance to the matching performance.
% Hence, the model will learn to assign higher weights to the intent.
Generally, we iteratively mask each intent and calculate the matching loss in each case.
If the loss increases by a large margin, it means that the masked intent is of great importance to the matching performance.
Hence, the model will learn to assign higher weights to the intent.

The corresponding process is shown in Figure~\ref{fig:overview}, where the loss before and after masking is denoted by the red and white bars, respectively.
Specifically, we keep the denotation of the original matching loss without the mask as $L_{match}$, and we denote the loss after the mask as $L_{new}$.
Then, the masking loss is  defined as:
\begin{align}
    \mathcal{L}_{mask}=\sum_{j=1}^{2c}\left\| \exp(\mathcal{L}_{new}-\mathcal{L}_{match}),\ \beta_j\right\|_2,
\end{align}
where $=\| \cdot \|_2$ is the L2 norm of the vector.

We use the overall loss $\mathcal{L}$ to optimize the parameters of the end-to-end framework:
\begin{align}
    \mathcal{L}=\mathcal{L}_{match}+\mathcal{L}_{dis}+\mathcal{L}_{KL}+\mathcal{L}_{mask}.
\end{align}

\section{Experiments}

% \subsection{Offline Experiments}

\subsection{Datasets}

\begin{table}[h]
\centering
    \small
    \caption{Dataset statistics.}
    \begin{tabular}{@{}c|l|l@{}}
      \toprule
      % \multirow{5}{*}{Amazon} & number of queries & 97,345 \\
      \multirow{4}{*}{Amazon} & number of query-item pairs & 1,818,825 \\
      & training dataset size & 27,757 \\
      & valid dataset size & 139,306\\
      & test dataset size & 425,762 \\
      \midrule
      \multirow{4}{*}{\makecell[c]{Alipay \\ Query Rewriting}} & number of rewritten query pairs & 32,604 \\
      & training dataset size & 1,253,757 \\
      & valid dataset size & 2,347\\
      & test dataset size & 2,500 \\
      \midrule
      \multirow{4}{*}{\makecell[c]{Alipay \\ Query-Item Matching}} & number of query-item pairs & 730,027 \\
      & training dataset size & 710,027 \\
      & valid dataset size & 10,000\\
      & test dataset size & 10,000 \\
      \bottomrule
    \end{tabular}
    \label{tab:dataset}
\end{table}

\begin{table*}[t]
    \centering
    \small
    \caption{Offline performance of MIM and baselines on Amazon and Alipay datasets. 
    Numbers in \textbf{bold} mean that the improvement to the best baseline is statistically significant (a two-tailed paired t-test with p-value \textless 0.01).}
    \begin{tabular}{@{}l|ccc|ccc|ccc@{}}
      \toprule
      & \multicolumn{3}{c|}{Amazon} & \multicolumn{3}{c|}{Alipay Query Rewriting} &  \multicolumn{3}{c}{Alipay Query-Item Matching} \\
      \midrule
      & Accuracy & AUC & F1 & Accuracy & AUC & F1 & Accuracy & AUC & F1\\
      \midrule
      % \multicolumn{4}{l}{\textit{Naive text-matching methods}} & \\
      BERT~\cite{kenton2019bert} & 0.7315 & 0.7767 & 0.8145 & 0.8560 & 0.9092 & 0.6944 & 0.8051 & 0.8722 & 0.8127\\
      SimCSE~\cite{gao2021simcse} & 0.7423 & 0.7882 & 0.8213 & 0.8577 & 0.9104 & 0.7003 & 0.8147 & 0.8736 & 0.8143\\
      RankCSE~\cite{liu2023rankcse} & 0.7495 & 0.8071 & 0.8247 & 0.8649 & 0.9192 & 0.7218 & 0.8154 & 0.8785 & 0.8166\\
      \midrule
      % \multicolumn{4}{l}{\textit{Multi-view text-matching methods}} & \\
      MVR~\cite{zhang2022multi} & 0.7492 & 0.8036 & 0.8275 & 0.8622 & 0.9137 & 0.7151 & 0.8137 & 0.8773 & 0.8182 \\
      MADRAL~\cite{kong2022multi} & 0.7624 & 0.8177 & 0.8229 & 0.8670 & 0.9188 & 0.7332 & 0.8142 & 0.8758 & 0.8184 \\
      \midrule
      % \multicolumn{4}{l}{\textit{Attributes-aware text-matching methods}} & \\
      BERT-concat & 0.7434 & 0.7941 & 0.8238 & 0.8578 & 0.9197 & 0.7102 & 0.8168 & 0.8764 & 0.8195\\
      AGREE~\cite{shan2023beyond} & 0.7681 & 0.8193 & 0.8282 & 0.8624 & 0.9129 & 0.7257 & 0.8169 & 0.8797 & 0.8213 \\
      Machop~\cite{wang2022machop} & 0.7705 & 0.8314 & 0.8303 & 0.8695 & 0.9136 & 0.7384 & 0.8213 & 0.8903 & 0.8340 \\
      DCMatch~\cite{zou2022divide} & 0.7732 & 0.8356 & 0.8343 & 0.8700 & 0.9208 & 0.7399 & 0.8202 & 0.8847 & 0.8366 \\
      \midrule
      MIM & \textbf{0.7813} & \textbf{0.8449} & \textbf{0.8413} & \textbf{0.8792} & \textbf{0.9324} & \textbf{0.7506} & \textbf{0.8343} & \textbf{0.9012} & \textbf{0.8431} \\
      \bottomrule
    \end{tabular}
    \label{tab:baselines}
  \end{table*}
  
In this paper, we conduct offline experiments on three real-world datasets as shown in Table~\ref{tab:dataset}, including a public benchmark Amazon~\cite{reddy2022shopping} and two large-scale industrial datasets that we collected from Alipay App.

The Amazon dataset is a query-item matching dataset.
It provides a list of up to 40 potentially relevant items for each query.
% , together with the relevance annotation.
Here we utilize the English part of the large version for the query-item matching task.
For the input query, we extract key phrases from it by TextRank~\cite{mihalcea2004textrank} as the attributes.
As for the item, its attributes include description, bullet point, brand, color, and source.
Following the official setting, all matching labels are split into training and test data.
% The training dataset contains 74,888 queries and 1,393,063 labels, and the test dataset consists of 22,458 queries and 425,762 labels.
We split 10\% of the training data as the valid dataset.

The Alipay Query Rewriting dataset is collected from the online Alipay searching platform, where we gather the consecutive exposure and click logs for four weeks and summarize the high-frequency queries.
We randomly sample a set of queries as inputs.
The corresponding candidate rewriting queries are constructed by human annotators. 
% Together, we obtain 32,604 query-rewritten query pairs.
% The pairs are split into a training dataset with 27,757 query pairs, a valid dataset with 2,347 query pairs and a test dataset with 2,500 query pairs.
% There are three attributes included in each query, i.e., entity, location, and category, where entity indicates the identified entity contained in the query, location is the geographic information contained in the query including countries, provinces, cities, and regions, and category identifies the functional category belonging to the query, such as transportation, technology, and medical health.
Each query includes three attributes: entity (the identified entity), location (geographic information such as countries, provinces, cities, and regions), and category (functional categories like transportation, technology, and medical health).
In the construction process for attributes, we asked annotators to exhaustively label a list of possible entities and locations.
Then, the entity and location attributes are obtained by exact string matching with the labeled list.
For the category classification, we use a pretrained model to match the queries to 24 pre-defined categories.
% For the category classification, we ask human annotators to match the queries to 24 pre-defined categories.
% We use the collected 282,770 labels to train a BERT model, which is then used to classify the 32,604 queries in the dataset.
% For the category classification, we ask human annotators to classify the queries into 24 pre-defined categories, and 282,770 collected labels are used to train a BERT model, which is then used to classify the 32,604 queries in the dataset.
% The rich attributes in this large-scale dataset can help MIM better recognize the query latent intent that users are interested in, and provide a better search experience.

The Alipay Query-Item Matching dataset is also constructed based on the high-frequency queries obtained above.
We obtain potentially relevant items for each query based on user click behavior log data, and all retrieved items are manually annotated to indicate whether they are relevant to the query.
The extraction of attributes from queries is consistent with the Query Rewriting dataset.
As for the items, we utilize structured information filled out by users as attributes, including keyword, category, functionality, and brand.

\subsection{Baselines.}
We compared our proposed model against naive text-matching methods, multi-view methods, and attribute-aware methods.

The naive text-matching baselines include:
$\bullet$ \textbf{BERT}~\cite{kenton2019bert}: the most commonly used pre-trained language model for text matching.
% , which learns deep bidirectional representations by joint conditioning on both left and right context in all layers.
$\bullet$ \textbf{SimCSE}~\cite{gao2021simcse}: a contrastive learning framework that advances  sentence embeddings.
$\bullet$ \textbf{RankCSE}~\cite{liu2023rankcse}: an unsupervised sentence representation learning method, which incorporates ranking consistency and ranking distillation with contrastive learning.
% The candidate query with the highest similarity with the input query is selected as the rewritten query.

Next, we compared with  baselines that learn multi-view representations for attributes which is similar to our approach:
$\bullet$ \textbf{MVR}~\cite{zhang2022multi}: extract multi-view embeddings for one document and align with different queries.
In this paper, we adapt their method to obtain multi-view representations by concatenating text input and attributes.
$\bullet$ \textbf{MADRAL}~\cite{kong2022multi}: introduces multi-attribute prediction task along with the main matching task to fuse the learned attribute information with the text semantic meanings.
% We learn multi-aspect from the input following the same setting.

Finally, we compared with baselines that also introduce attributes as inputs:
$\bullet$ \textbf{BERT-concat}: a baseline that directly concatenates the input text with attributes as BERT input.
$\bullet$ \textbf{Machop}~\cite{wang2022machop} employs attributes as additional knowledge to guide the model to focus on key-matching information.
% $\bullet$ \textbf{Machop}~\cite{wang2022machop} casts the generalized entity matching problem as sequence pair classification so as to utilize the language understanding capability of language models.
% It takes attributes as additional knowledge to guide the model to focus on the key-matching information.
$\bullet$ \textbf{DCMatch}~\cite{zou2022divide} designs multiple supervised tasks to learn different granularities of attribute information for matching.
$\bullet$ \textbf{AGREE}~\cite{shan2023beyond} introduces an attribute fusion layer in the item side to incorporate the most relevant item features into representations.

\subsection{Implementation Details}
We implemented our experiments in Pytorch on an NVIDIA A100 GPU.
For our model and all baselines, we followed the same setting as described below.
We set the truncation length for the Amazon dataset as 512 and 128 for the Alipay dataset.
We set the intent number to three by default.
% The model is fine-tuned based on BERT-base with 12 layers and 768 hidden dimensions, and the other parameters are all initialized randomly using a Gaussian distribution.
The model was fine-tuned based on BERT-base with 12 layers and 768 hidden dimensions, and the other parameters were all initialized randomly using a zero-mean Gaussian distribution with a std of 0.01.
Experiments were performed with a batch size of 256.
% Based on the observation that the model performs best with three intents, we set the number of intents to three by default.
We used Adam optimizer~\cite{kingma2014adam} as our optimizing algorithm, with a range of $[-1,1]$ gradient clipping during training.
We selected the 5 best checkpoints based on performance on the validation set and reported averaged results on the test set.

\subsection{Evaluation Metrics}
For offline evaluation, we use Accuracy, AUC, and F1 metrics, which have been widely adopted for candidate matching:
$\bullet$ \textbf{Accuracy} calculates the proportion of correct predictions among all prediction results.
% , which is widely used for classification.
$\bullet$ \textbf{AUC} calculates the area under the ROC curve, which can better adapt to the phenomenon of class imbalance.
$\bullet$ \textbf{F1 score} balances both recall and precision, and leads to a better judgment about the performance of the classifier.
% takes both recall and precision into account, balances the two scores, and leads to a better judgment about the performance of the classifier.
% The higher the F1 score, the higher the balance between precision and recall achieves.

For online evaluation, we select a range of core commercial metrics, including the click-through-rate per page view (pvCTR) and Relevance Score metrics consisting of good case rate (GR) and bad case rate (BR).
% \indent $\bullet$ \textbf{pvCTR} calculates the number of clicked times divided by the number of searched times.
$\bullet$ \textbf{pvCTR} calculates the average number of clicked times per search action.
$\bullet$ \textbf{GR} is used to evaluate the proportion of accurate paraphrasing cases, or the proportion of items that fully meet the user's query requirements, whose growth reflects the benefits actually brought to the Alipay APP.
$\bullet$ \textbf{BR}, corresponding to GR, calculates the proportion of bad results. Its decrease indicates that the model has actually solved existing problems.

% \indent $\bullet$ \textbf{GSB} measures the percentage of better results returned by the rewritten queries.
% Annotators are given the end-to-end search results returned by the original queries and rewritten queries, and they are asked to label whether the results of the rewritten queries outperform the original queries.

% \indent $\bullet$ Relevance scores between the matched queries are evaluated by human annotators.
% Here, we show the original query and the candidate rewritten queries selected by our model and baselines to the annotators, and ask them to label the proper pairs that should be rewritten.
% The scores include \textbf{GR}, \textbf{MR}, and \textbf{BR}, which are the percentage of good, medium, and bad cases.

\subsection{Offline Performance
}
\label{sec:analysis}

% \textbf{Overall Performance.}
Table~\ref{tab:baselines} shows the experimental results in terms of Accuracy, AUC, and F1 score on the three datasets. 
% Here, we can make several observations.
Firstly, the supervised learning of text matching models performs better than that of representation learning models.
This is because, through task design, the model can better learn which information in the input data should contribute more to the matching task, while ignoring the parts that may have meaning but do not contribute much to judging text matching.
For example, even if the recent RankCSE model shows excellent performance on embedding learning tasks, it still does not perform as well as several baselines designed for matching, which means sufficient interaction is decisive in the fusion of information, and its effectiveness is significant in matching tasks.
Secondly, models with attributes show better performance, suggesting that the attribute information is of great significance in the text-matching task.
In comparison with the BERT-concat, Machop, and DCMatch show better ability in candidate matching. 
This may be mainly attributed to the fact that the Machop and DCMatch can better utilize attributes.
AGREE and MADRAL both have strong performance by using attributes as additional supervision signals.
% Second, For baselines with cross-attention strategy, i.e., BERT-concat, Machop, and MIM, their performances are significantly improved. 
% On the contrary, SimCSE, COIN, CARCA, and MVR only have bi-directional attention, and their performance is relatively weak.
% This mainly demonstrates that the full flow of information throughout the two inputs is necessary.
% Sufficient interaction is decisive in the fusion of information, and its effectiveness is significant in matching tasks.
Thirdly, we find that multi-aspect modeling is effective in various directions.
For example, MVR is a method with a max-pooling operation on multi-view representations, and Machop has multi-attribute structure-aware pooling.
They both achieve good performance compared with other baselines.
Finally, MIM achieves the best performance across all the datasets and different metrics.
Specifically, MIM outperforms DCMatch by 0.0091 AUC score on Amazon, 0.0116 score on Alipay Query Rewriting datasets, and 0.0165 score on Alipay Query-Item Matching tasks, respectively.
All pairwise comparisons among systems are statistically significant using the paired student t-test for significance at $\alpha = 0.01$.
The above observation demonstrates that our method of extracting intents makes better use of the attributes, and our designed intent alignment mechanism improves the text-matching performance.

\begin{table}[t]
    \centering
    \small
    \caption{The performance improvements on two scenes of Alipay App Search with online A/B test.}
    \begin{tabular}{@{}l|ccc|ccc@{}}
      \toprule
      & \multicolumn{3}{c|}{Alipay Query Rewriting} &  \multicolumn{3}{c}{Alipay Query-Item Matching} \\
      Method & pvCTR & GR & BR & pvCTR & GR & BR \\
      \midrule
      online & +0.0 & +0.0 & -0.0& +0.0 & +0.0 & -0.0 \\
      MIM & \textbf{+0.11\%} & \textbf{+11.95\%} & \textbf{-13.43\%} & \textbf{+0.84\%} & \textbf{+2.32\%} & \textbf{-1.15\%} \\
      \bottomrule
    \end{tabular}
    \label{tab:online_evaluation}
  \end{table}

\begin{table}
  \centering
\small
\caption{Ablation study of MIM on Amazon dataset. Numbers in \textbf{bold} indicate that the improvement over the best baseline is statistically significant.}
\begin{tabular}{@{}lccc@{}}
  \toprule
  & Accuracy & AUC & F1 \\
  \midrule
  MIM & \textbf{0.7813} & \textbf{0.8449} & \textbf{0.8413} \\
  \midrule
  w/o gate score & 0.7763 & 0.8429 & 0.8402 \\
  w/o $\mathcal{L}_{KL}$ & 0.7741 & 0.8403 & 0.8384 \\
  w/o $\mathcal{L}_{dis}$ & 0.7744 & 0.8392 & 0.8387 \\
  w/o multi-intent & 0.7664 & 0.8254 & 0.8332 \\
  w/o mask task & 0.7780 & 0.8435 & 0.8397 \\
  \bottomrule
\end{tabular}
\label{tab:ablation_study}
\end{table}

% \begin{table}[t]
%     \centering
%     \small
%     \caption{The performance improvements on two scenes of Alipay App Search with online A/B test.}
%     \begin{tabular}{@{}l|ccc@{}}
%       \toprule
%       Method & pvCTR & GR & BR  \\
%       \midrule
%       online & +0.0 & +0.0 & -0.0 \\
%       MIM & \textbf{+0.49\%} & \textbf{+11.95\%} & \textbf{-13.43\%}  \\
%       \bottomrule
%     \end{tabular}
%     \label{tab:online_evaluation}
%   \end{table}

\subsection{Online Experiments}
Besides the offline experiments, we also conduct online A/B tests by deploying MIM in the query rewriting and query-item relevance matching of the Alipay searching system for seven days. 
% In the control group, the matching strategy deployed in our current online system, i.e., the BERT-concat model, is taken as the baseline. 
In the control group, the matching strategy deployed in our current online system follows a similar cross-encoder structure.
% The BERT-concat is chosen because it is the model we are currently using online.
In the test group, the proposed MIM is deployed to serve the candidate rewritten query matching and query-item relevance matching, respectively.
The same recall and ranking strategies are used for a fair comparison, and the averaged A/B test results are reported in Table~\ref{tab:online_evaluation}. 
Note that the results are reported with relative improvements. 
Here, it can be observed that the proposed MIM provides better performance on all online score metrics.
Specifically, it outperforms the current system in terms of the good case rate and bad case rate by a large margin, which indicates that MIM enhances the user's search experience.
Our MIM also achieves a relative improvement of 0.11\% and 0.84\% in terms of the pvCTR on both scenarios respectively.
% Our MIM also has a higher pvCTR score compared with the baseline.
% Note that by improving the accuracy of query rewriting, the number of retrieved results is reduced.
% Nevertheless, the pvCTR score still improves, which shows that our model improves user satisfaction by providing more accurate results.
% The fluctuation of pvCTR shows that narrowing retrieval results to 32.3\%, caused by improving the rewrite relevance, slightly improves the overall efficiency with enhancing rewrite accuracy.
% does not have a negative impact on the global efficiency index.
Therefore, we can conclude that MIM generates better search results for millions of users in Alipay.

\section{ANALYSIS AND DISCUSSION}

\subsection{Ablation Study.}
We perform an ablation study on the test set to investigate the influence of different modules in our proposed MIM model. 
Modules are tested in four ways: (1) We remove the gate score in the attribute-aware encoder, and use the original self-attention mechanism instead; (2) we remove the kl loss $\mathcal{L}_{KL}$ which is used to align the distribution of intents on both sides; (3) we remove the distribution loss $\mathcal{L}_{dis}$ that models multiple intents distribution around the text representation; (4) we remove the multi-intent modeling module completely and directly feed the output from the encoder to the matching module; (5) we remove the self-supervised mask task in the intent-aware matching.

% To further investigate the impact of different modules (i.e., mask loss, intent loss, multi-intent modeling, and weightes on attribute) in our proposed MIM, we conduct the ablation study on the Amazon dataset. 
% To verify each component’s impact, we disable one component each time and keep the other parts unchanged.
% Table~\ref{tab:ablation_study} reports the performance of these variants. 
% According to the result, it is obvious that MIM obtains the best performance across different metrics, confirming the validity of these components.

Table~\ref{tab:ablation_study} presents the results on the Amazon dataset.
First, the F1 score drops by 0.0011 after removing the scaled self-attention mechanism, which indicates that the attribute content is of great importance to the attention process.
Then, the AUC score drops by 0.0046 and 0.0057 after the $\mathcal{L}_{KL}$ loss and $\mathcal{L}_{dis}$ are removed, respectively. 
This shows that restrictions on the learned intents do help identify matching aspects of two inputs.
The F1 score drops by 0.0081 after the multi-intent modeling is removed. 
It denotes establishing the intent modeling in the matching process is necessary to improve the performance of text matching. 
Finally, we find that the Accuracy score drops by 0.0033 after the mask loss is removed. 
This means that the additional self-supervised task helps the matching model to benefit from proper intent weights.

\begin{figure} 
\centering 
\subfigure[]{ 
  \includegraphics[width=0.47\linewidth]{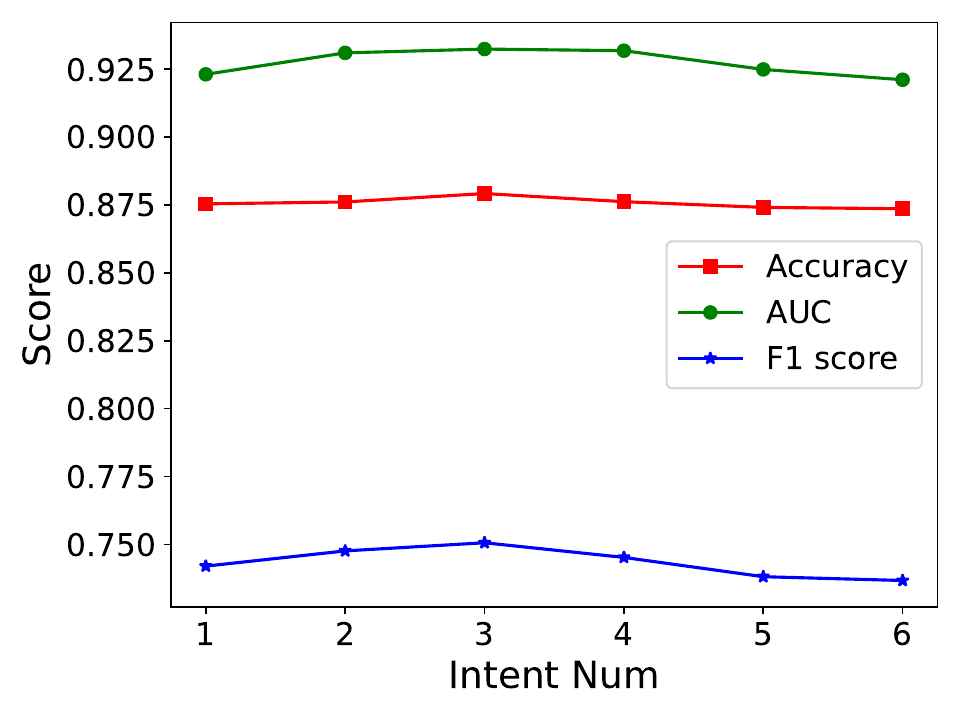}
}
\subfigure[]{ 
  \includegraphics[width=0.47\linewidth]{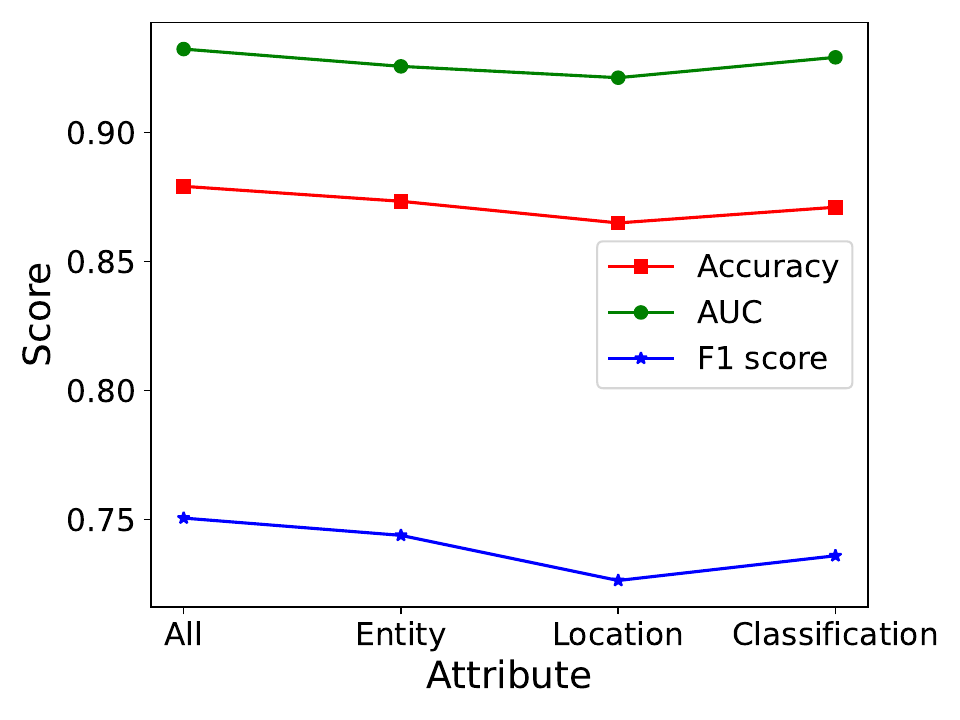}
}
\caption{(a) Relationship between number of intent and Accuracy, AUC and F1 score on Alipay test dataset. (b) Performance of MIM after removing different attributes in Alipay dataset.} 
\label{fig:subfig}
\end{figure}

\textbf{Number of Intents.}
We change the intents number in the Aliapy Query Rewriting dataset from 1 to 6, and report the Accuracy, AUC, and F1 scores as shown in Figure~\ref{fig:subfig}(a). 
It can be observed that the metric scores increase with augmenting the number of intents, to begin with.
This is intuitive since one or two intents cannot cover the diverse semantic meanings of the inputs.
After reaching the upper limit, it begins to drop.
This might be because the differences between intents become trivial, and the model cannot utilize them well to conduct matching.
Generally, our model maintains a high performance across different intent numbers.

\textbf{Impact of Each Attribute.}
We are also interested in each attribute's contribution, i.e., which attribute is most important for the text-matching task.
To illustrate the effect of attribute representations in MIM, we investigate how MIM performs after removing each attribute on the Alipay Query Rewriting dataset.
We plot the corresponding Accuracy, AUC, and F1 scores in Figure~\ref{fig:subfig}(b).
It is noted that for the Alipay searching scenario, the location area plays an important role.
When we remove this attribute, the performance of MIM drops the most.

\begin{figure}[t]
    \centering
    \includegraphics[scale=0.2]{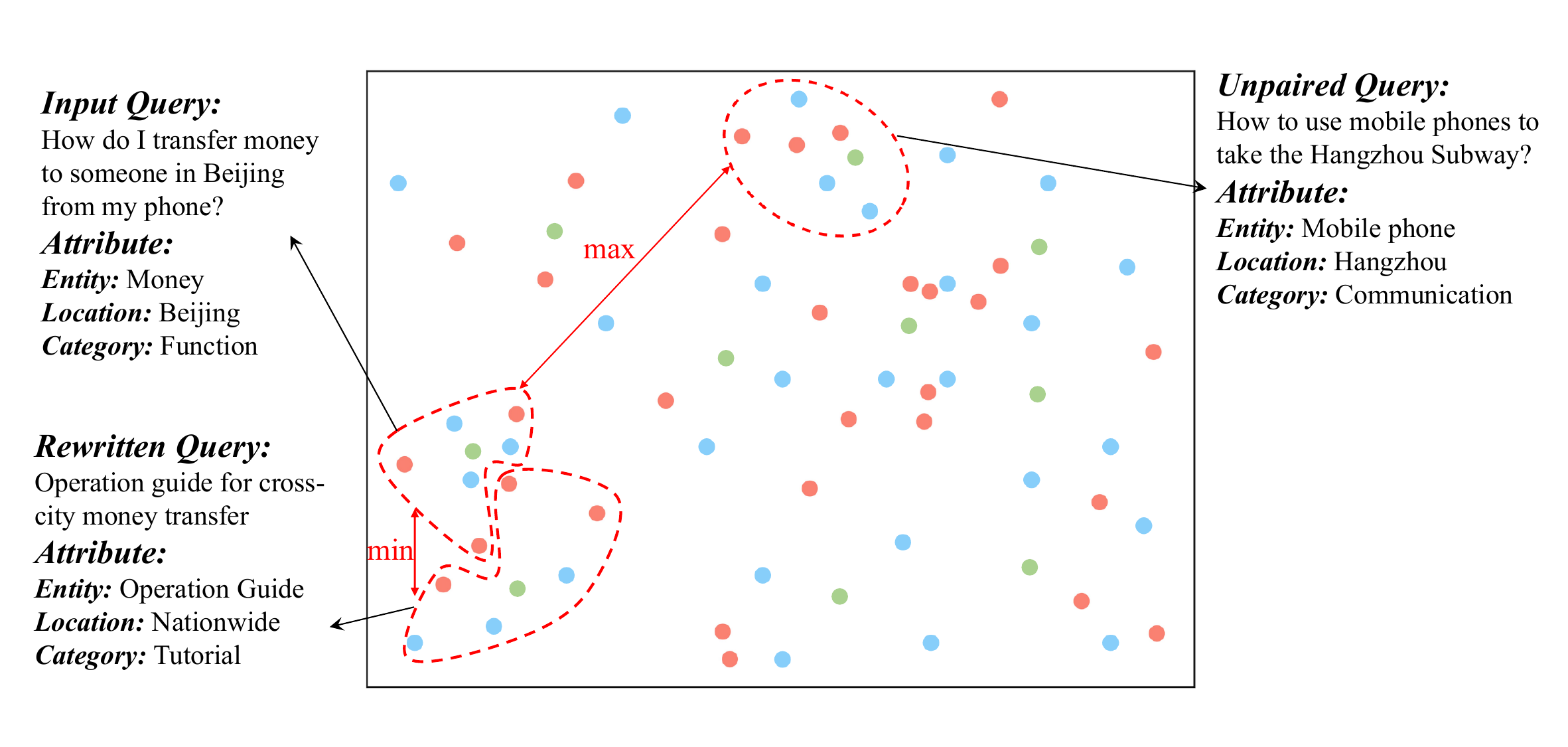}
    \caption{
        Visualization of intent distribution with attributes, where ten cases are sampled from the test set randomly. The \textcolor[RGB]{255,99,71}{red nodes}, \textcolor[RGB]{135,206,235}{blue nodes}, and \textcolor[RGB]{60,179,113}{green nodes} represent the \textcolor[RGB]{255,99,71}{intents}, \textcolor[RGB]{135,206,235}{attributes}, and \textcolor[RGB]{60,179,113}{queries}, respectively.}
    \label{fig:intent_dis}
\end{figure}

\begin{table}
    \centering
    \scriptsize
%   \centering
% \small
\caption{Case study (truncated) for Alipay Query Rewriting Task.
    $\bullet$ denotes attributes.
    The highlighted \hlc[pink]{pink-colored words} indicate the aligned needs, and the \hlc[cyan!50]{blue-colored words} show the misaligned information.}
\begin{tabular}{@{}l|l@{}}
  \toprule
  \multicolumn{2}{c}{Query Rewriting \#1} \\
  \midrule
  \textit{Input Query:} & $\bullet$ Entity: Electronic driving license \\
  How to check \hlc[cyan!50]{violation records} of   &
      $\bullet$ Location area: Nationwide  \\ 
    \hlc[pink]{electronic driving license}? & $\bullet$ Category: Electronic certificate inquiry  \\
      \midrule
\textit{Relevant Rewritten Query Selected by MIM:} & $\bullet$ Entity: Traffic management\\
      Traffic management 12123 &
$\bullet$ Location area: Nationwide  \\ 
      & $\bullet$ Category: \hlc[pink]{Motor vehicle management}  \\
      \midrule
    \textit{Relevant Rewritten Query Selected by DCMatch:} & $\bullet$ Entity: \hlc[pink]{digital ID}  \\
      How can I \hlc[cyan!50]{capture a photo} for my \hlc[pink]{digital ID}? & $\bullet$ Location area: Nationwide \\ 
      & $\bullet$ Category: Domestic services  \\
      \bottomrule
      
      \multicolumn{2}{c}{Query Rewriting \#2} \\
  \midrule
  \textit{Input Query:} & $\bullet$ Entity: \hlc[pink]{medical insurance} \\
  What is the \hlc[pink]{medical insurance} \hlc[cyan!50]{electronic }&  $\bullet$ Location area: Beijing  \\ 
      \hlc[cyan!50]{certificate} in Beijing? & $\bullet$ Category: Medical care assurance  \\
      \midrule
\textit{Relevant Rewritten Query Selected by MIM:} & $\bullet$ Entity: Insurance card \\
      Beijing citizen \hlc[pink]{electronic health insurance} card &
       $\bullet$ Location area: Beijing   \\ 
      & $\bullet$ Category: Medical care assurance   \\
      \midrule
      \textit{Relevant Rewritten Query Selected by DCMatch:} & $\bullet$ Entity: \hlc[pink]{Medical insurance} \\
      What is the process for \hlc[cyan!50]{purchasing medicine}  & $\bullet$ Location area: Beijing   \\ 
       \hlc[cyan!50]{online} using \hlc[pink]{medical insurance} in Beijing? & $\bullet$ Category: Medical health   \\
      
  \bottomrule
\end{tabular}
\label{tab:case_study}
\end{table}

\subsection{Visualization of Intent Distribution with Attributes}
Since our extracted intents are represented by vectors, it is hard to directly obtain their semantic meanings.
Herein, we randomly sample ten cases and project their query, attribute, and intent representations into a two-dimension space with t-SNE \citep{maaten2008visualizing}.
As shown in Figure~\ref{fig:intent_dis}, the intents of matched queries or items are close to each other, while the unpaired representations are scattered in the space.
The extracted intents are evenly distributed and surround the query. The proximity of the intent and attributes of the same query indicates that the intent effectively summarizes the information contained in the attributes.

\subsection{Case Study.}
In Table~\ref{tab:case_study}, we randomly select two cases from the Alipay Query Rewriting datasets, where we show the input queries, attributes of the queries and items, and output results returned by our MIM and the best baseline DCMatch.
The highlighted pink-colored words indicate the aligned needs, and the blue-colored words show the misaligned information.
It can be seen that our MIM successfully captures the intents of checking records and searching for electronic cards in the original query, and the input query is rewritten to a more simple, condensed, and accurate query.
% We show more cases in Appendix.
% For the query-item matching task, the relevant item returned by our MIM meets the input query's needs.
% However, the item selected by Machop focuses on the information ``supplement without tea leaves'' and ignores the keyword ``resveratrol'', which leads to a irrelevant result.

\section{Conclusion}
\label{sec:conclusion}
In this paper, we propose a multi-intent attribute-aware text matching model, which matches queries to rewrite queries or items by attending to attributes and extracting intents.
To address the problem that multiple intents collapse into similar representations, we propose a distribution loss to increase the diversity of the intents related to the query text.
We also come up with a divergence loss to better pair the learned intents from both sides.
We finally employ an intent-mask self-supervision task to decide the importance of each intent and incorporate them to output the final matching result.
% Extensive experiments demonstrate the effectiveness of the proposed MIM and we are looking forward to expanding textual attribute-aware matching to multimodal scenario in the future.
Extensive experiments demonstrate the effectiveness of the proposed MIM, which currently serves tens of millions of users on Alipay app online.
In the future, we aim to expand textual attribute-aware matching to multimodal attribute-aware matching.

\end{CJK*}

\bibliographystyle{ACM-Reference-Format}
\balance
\bibliography{www2021}

  \end{document}

\begin{CJK*}{UTF8}{gkai}

\section{Appendix}
\begin{table*}[h]
\newcolumntype{I}{!{\vrule width 1pt}}
    \centering
    \small
    \caption{Case study for query rewriting and query-item matching tasks.
    % For the query rewriting task, the input query is rewritten to a more simple, condensed, and accurate query. 
    $\bullet$ denotes attributes.
    }
    \begin{tabular}{@{}lIl@{}}
      \toprule

      \multicolumn{1}{cI}{Query Rewriting \#1} & \multicolumn{1}{c}{Query-Item Matching}\\
      \cline{1-2}
      \textit{Input Query:} & \textit{Input Query:}\\
      How to check violation records of electronic driving license? & \#1 rated resveratrol supplement without tea leaves \\
      \quad $\bullet$ Entity: Electronic driving license  & \quad $\bullet$ Keywords: resveratrol supplement \\
      \cline{2-2}
      \quad $\bullet$ Location area: Nationwide  & \textit{Relevant Item Selected by MIM:}\\ 
      \quad $\bullet$ Category: Electronic certificate inquiry  & \multirow{2}{9cm}{Reserveage, Resveratrol 500 mg, Antioxidant Supplement for Heart and Cellular Health, Supports Healthy Aging, Paleo, Keto, 60 Capsules}\\
      \cline{1-1}
      \textit{Relevant Rewritten Query Selected by MIM:}  & \\
      Traffic management 12123 &  \multirow{2}{9cm}{\quad $\bullet$ Bullet point: Age-defying resveratrol, Vegan, paleo-friendly, 1 veggie capsule, Triactiv technology}\\
      \quad $\bullet$ Entity: Traffic management  & \\
      \quad $\bullet$ Location area: Nationwide  & \quad $\bullet$ Brand: Reserveage Nutrition\\ 
      \quad $\bullet$ Category: Motor vehicle management  & \quad $\bullet$ Color: Default \\
      \cline{1-1}
      \textit{Relevant Rewritten Query Selected by Machop:} & \quad $\bullet$ Source: Amazon \\
      Where can I take a photo for my E-document? &  \multirow{11}{9cm}{\quad $\bullet$ 
      Description: Age-defying resveratrol: An advanced, triple blend formula made to support healthy heart function, cellular health and to help activate the "longevity gene".
Vegan, paleo-friendly: Our Pro-Longevity Factors Blend includes natural ingredients like organic French and Muscadine red wine grapes and Wildcrafted Japanese Knotweed.
1 veggie capsule: Taken once daily delivers 500 mg of our proprietary Resveratrol; Can be taken on an empty stomach or with food; Free of gluten, soy, sugar and preservatives.
Triactiv technology: Made from 3 blends of trans-Resveratrol for a steady release of antioxidant protection lasting up to 4 hours.
With Reserveage Nutrition Seal of Authenticity: Ensures you are getting an authentic product from the manufacturer and can purchase with confidence.}

\\
      \quad $\bullet$ Entity: E-document  &  \\ 
      \quad $\bullet$ Location area: Nationwide  & \\ 
      \quad $\bullet$ Category: Domestic services  &  \\
    \cline{1-1}

      \multicolumn{1}{cI}{Query Rewriting \#2} & \\
      \cline{1-1}
      \textit{Input Query:} & \\
      What is the medical insurance electronic certificate in Beijing? & \\
      \quad $\bullet$ Entity: Medical insurance  & \\
      \quad $\bullet$ Location area: Beijing  &  \\ 
      \quad $\bullet$ Category: People's livelihood and government affairs  & \\ 
      \cline{1-1}
      \textit{Relevant Rewritten Query Selected by MIM:} & \\
      \cline{2-2}
      Beijing citizen electronic health insurance card & \textit{Relevant Item Selected by Machop:}
      \\
      \quad $\bullet$ Entity: Insurance card  & Turmeric Curcumin Supplement with Ginger \& Apple Cider Vinegar\\ 
      \quad $\bullet$ Location area: Beijing  &  \multirow{2}{9cm}{\quad $\bullet$ Bullet point: Trusted Formula: Turmeric and ginger are two of nature's most powerful ingredients.
} \\ 
      \quad $\bullet$ Category: Medical care assurance  &  \\
      \cline{1-1}
      \textit{Relevant Rewritten Query Selected by Machop:} & \quad $\bullet$ Brand: NATURE'S BASE\\
      How to buy medicine online with medical insurance in Beijing? &  \quad $\bullet$ Color: None\\
      \quad $\bullet$ Entity: Medical insurance  & \quad $\bullet$ Source: Amazon \\ 
      \quad $\bullet$ Location area: Beijing  &  \multirow{14}{9cm}{\quad $\bullet$ Description: Nature's Base Turmeric Curcumin with Ginger and Apple Cider Vinegar, 95\% Curcuminoids, Tumeric Supplements, Occasional Joint Relief, Inflammatory Response, Natural Plant Based Anti-Oxidant Properties. Do not exceed the recommended dose. Pregnant or nursing mothers, children under the age of 18, and individuals with a known medical condition should consult a physician before using this or any other dietary supplement. DO NOT use if safety seal is broken or missing. Store in a cool, dry place. Before beginning daily use, take 1 capsule and wait 24 hours to rule out any possibility of any potential side effects. Warning: Please note that in some rare cases taking supplements may cause upset stomach, heartburn, bloating, constipation, fatigue, headache, dizziness, nausea, vomiting, diarrhea, reactions including rash, itching, coughing, swelling, or sickness. If you experience these or any other adverse effects or other reactions, you should immediately discontinue use of the product and contact your physician.}\\ 
      \quad $\bullet$ Category: Medical health  &  \\
        &  \\  &  \\  &  \\  &  \\  &  \\  &  \\  &  \\  &  \\  &  \\  &  \\  &  \\  &  \\
      
      \bottomrule
    \end{tabular}
    \label{tab:entire_case}
  \end{table*}

% \section{Case Study}
We show the entire representative cases of Amazon and Alipay dataset for query rewriting and query-item matching tasks in Table~\ref{tab:entire_case}.

\end{CJK*}

% \bibliographystyle{ACM-Reference-Format}
% \bibliography{www2021}

% % \newpage
% \appendix